\documentclass[10pt,twocolumn]{article} 
\usepackage{simpleConference}
\usepackage{times}
\usepackage{graphicx}
\usepackage{amssymb}
\usepackage{gensymb}
\usepackage{amsmath}
\usepackage{url}
\usepackage{hyperref} 
\usepackage{breakurl}
\def\UrlBreaks{\do\/\do-}
\hypersetup{
colorlinks,
linkcolor=blue,
citecolor=blue,
filecolor=blue,
urlcolor=blue}
\usepackage[labelfont={bf},]{caption}
\usepackage[none]{hyphenat}

\usepackage{multirow}
\usepackage{color}

\usepackage{pdflscape}

\usepackage{tabularx}
\newcolumntype{Y}{>{\centering\arraybackslash}X}

\newcolumntype{A}[2]{%
    >{\minipage{\dimexpr#1\linewidth-2\tabcolsep-#2\arrayrulewidth\relax}\vspace\tabcolsep}%
    c<{\vspace\tabcolsep\endminipage}}

\usepackage[noadjust]{cite}

\usepackage{pifont}
\newcommand{\cmark}{\ding{51}}%
\newcommand{\xmark}{\ding{55}}%

\definecolor{mygreen}{rgb}{0,0.6,0}
\definecolor{mygray}{rgb}{0.5,0.5,0.5}
\definecolor{mymauve}{rgb}{0.58,0,0.82}

\newcommand{\longurl}[1]{%
    {\expandafter\def\expandafter\UrlBreaks\expandafter{\UrlBreaks\UrlOrds%
        \do\/\do\a\do\b\do\c\do\d\do\e\do\f%
        \do\g\do\h\do\i\do\j\do\k\do\l\do\m%
        \do\n\do\o\do\p\do\q\do\r\do\s\do\t%
        \do\u\do\v\do\w\do\x\do\y\do\z%
        \do\A\do\B\do\C\do\D\do\E\do\F\do\G%
        \do\H\do\I\do\J\do\K\do\L\do\M\do\N%
        \do\O\do\P\do\Q\do\R\do\S\do\T\do\U%
        \do\V\do\W\do\X\do\Y\do\Z}%
    \url{#1}}%
}

\usepackage[font=small]{caption}

\usepackage{etoolbox}
\makeatletter
\patchcmd\@combinedblfloats{\box\@outputbox}{\unvbox\@outputbox}{}{\errmessage{\noexpand patch failed}}
\makeatother

\newcommand\extraspace{3pt}
\newcommand\finalcapttxtstart{ Example applications of denoising autoencoders, kernels and MLPs. }
\newcommand\finalcapttxt{ More example applications of denoising autoencoders, kernels and MLPs. }

\begin{document}

\title{Autoencoders, Kernels, and Multilayer Perceptrons for Electron Micrograph Restoration and Compression}

\author{Jeffrey M. Ede \\
\\
j.m.ede@warwick.ac.uk  \\
}

\maketitle

\noindent \textbf{Abstract:} \textit{We present 14 autoencoders, 15 kernels and 14 multilayer perceptrons for electron micrograph restoration and compression. These have been trained for transmission electron microscopy (TEM), scanning transmission electron microscopy (STEM) and for both (TEM+STEM). TEM autoencoders have been trained for 1$\times$, 4$\times$, 16$\times$ and 64$\times$ compression, STEM autoencoders for 1$\times$, 4$\times$ and 16$\times$ compression and TEM+STEM autoencoders for 1$\times$, 2$\times$, 4$\times$, 8$\times$, 16$\times$, 32$\times$ and 64$\times$ compression. Kernels and multilayer perceptrons have been trained to approximate the denoising effect of the 4$\times$ compression autoencoders. Kernels for input sizes of 3, 5, 7, 11 and 15 have been fitted for TEM, STEM and TEM+STEM. TEM multilayer perceptrons have been trained with 1 hidden layer for input sizes of 3, 5 and 7 and with 2 hidden layers for input sizes of 5 and 7. STEM multilayer perceptrons have been trained with 1 hidden layer for input sizes of 3, 5 and 7. TEM+STEM multilayer perceptrons have been trained with 1 hidden layer for input sizes of 3, 5, 7 and 11 and with 2 hidden layers for input sizes of 3 and 7. Our code, example usage and pre-trained models are available at \url{https://github.com/Jeffrey-Ede/Denoising-Kernels-MLPs-Autoencoders}.}



\section{Introduction}
\noindent Many imaging modes in electron microscopy are limited by noise or generate large amounts of data that needs to be compressed. Nevertheless, there has been little development of electron microscopy specific denoising autoencoders, kernels and multilayer perceptrons (MLPs) for image restoration and compression. 

In this paper, we present a process to train noise removal kernels and MLPs for specific or general electron microscopy imaging modes. This is done by training denoising autoencoders for domains of interest and then training kernels and MLPs to approximate them. This has a number of advantages over just using an autoencoder, including:

\begin{figure}[tbp]
\centering
\includegraphics[width=0.57\columnwidth]{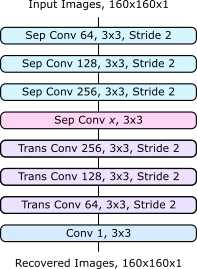}
\caption{ Denoising autoencoder with a 20$\times$20$\times$$x$ latent space.}
\label{autoencoder_architecture}
\end{figure}

\begin{table}[tbp]
\centering
\begin{tabularx}{\columnwidth}{lYYYYYYY}
\hline
              & \multicolumn{7}{c}{Latent Depth, $x$}    \\
Dataset       & 1 & 2  & 4   &8     & 16   &32    & 64       \\ \hline
TEM           & \cmark & \xmark & \cmark & \xmark & \cmark & \xmark & \cmark      \\
STEM          & \xmark & \xmark & \cmark & \xmark & \cmark & \xmark & \cmark \\
TEM+STEM      & \cmark & \cmark & \cmark & \cmark & \cmark & \cmark & \cmark \\
\hline
\end{tabularx}
\caption{ Autoencoders with 20$\times$20$\times$$x$ latent spaces trained for TEM, STEM and TEM+STEM. }
\label{autoencoders_table}
\end{table}

\begin{itemize}
\item Lower susceptibility to overfitting as a result of lower capacities.
\item Higher computational efficiency.
\item Improved resistance to checkerboard artefacts\cite{sugawara2018super}.
\end{itemize}

\noindent We have used our process to train autoencoders, kernels and MLPs for transmission electron microscopy (TEM), scanning transmission electron microscopy (STEM) and for both modes together (TEM+STEM). To be clear, in this paper TEM refers to all electron micrographs created through transmission; rather than rastering, and includes diffraction patterns.

Altogether, we present 14 autoencoders, 15 kernels and 14 MLPs for TEM, STEM and TEM+STEM restoration and compression. Autoencoders are presented in section~\ref{sec_autoencoders}. The kernels and MLPs trained to approximate 4$\times$ compression autoencoders are presented in section~\ref{sec_kernels+MLPs}. Performances are compared in section~\ref{sec_performance}. Finally, some of our kernels and example applications of each of the denoising autoencoders, kernels and MLPs are provided as an appendix.

\section{Autoencoders}\label{sec_autoencoders}

\noindent We have trained denoising autoencoders for TEM, STEM and TEM+STEM restoration and compression. These are simple convolution banks that compress images into latent spaces that are a fraction of their original size. These latent spaces can be stored as compressed data or decoded with a denoising effect.

\begin{figure}[tbp]
\centering
\includegraphics[trim={0.2cm 0.45cm 0 0},width=\columnwidth]{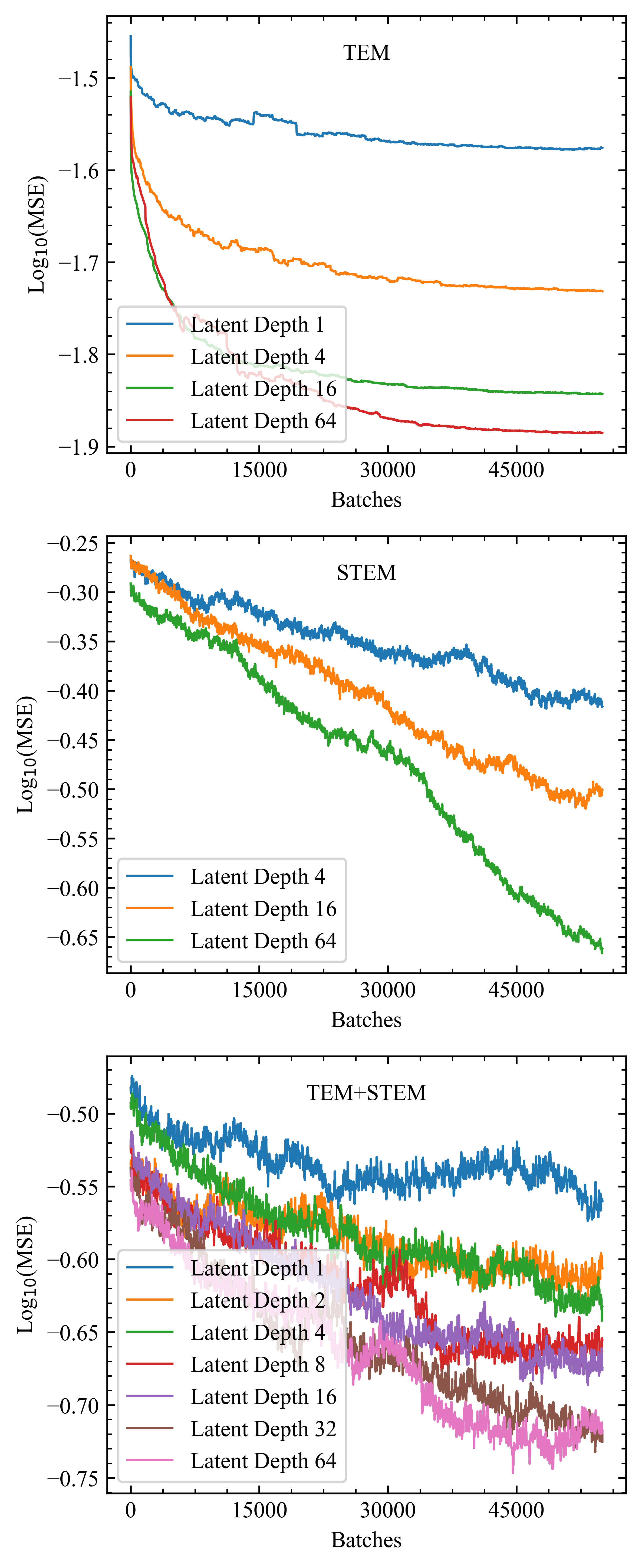}
\caption{ Autoencoder MSE learning curves for TEM, STEM and TEM+STEM. TEM MSEs have the lowest mean and variance, STEM MSEs have the highest mean and itermediate variance and TEM+STEM MSEs means are intermediate and have the highest variance. Learning curves are moving averages of 5000 batches. }
\label{autoenc_learning_curves}
\end{figure}

\subsection{Architecture}
\noindent Our autoencoder architecture is shown in fig.~\ref{autoencoder_architecture}. Each convolutional layer is followed by batch normalization, then ReLU\cite{nair2010rectified} activation. This is except for the last 2 layers: the second to last layer is not batch normalized and the last layer has no weights, biases, batch normalization or activation. 

All our autoencoders encode 20$\times$20$\times$$x$ latent spaces, with 64:$x$ compression ratios. Autoencoders trained for TEM, STEM and TEM+STEM with different latent depths are tabulated in table~\ref{autoencoders_table}.

\subsection{Learning Policy}
\noindent In this subsection, we present our autoencoder training hyperparameters and learning protocol for the learning curves shown in fig.~\ref{autoenc_learning_curves}.

\vspace{\extraspace}
\noindent \textbf{Loss metric:} Our autoencoders were trained to minimize the Huberised\cite{huber1964robust} mean squared error (MSE), $L_\mathrm{auto}$, between their outputs and inputs:

\begin{equation}
L_\mathrm{auto} = 
\begin{cases}
      \textit{MSE}, & \textit{MSE} < 1.0 \\
      \textit{MSE}^{\frac{1}{2}}, & \textit{MSE} \ge 1.0
\end{cases}
\end{equation}

The loss is Huberized to prevent autoencoders from being too disturbed by large errors in the early stages of training. Otherwise, it did not significantly affect training as most MSEs were of the order $10^{-1}$ or less.

To make all the 160$\times$160 micrograph crops in our training datasets of similar importance, their minima were subtracted and they were divided by their means. Dividing by means put differences between outputs and inputs on the same scale so that MSEs would be a meaningful loss metric. Minima subtraction was applied to increase the similarity of input intensity distributions, making it easier for the autoencoders to learn. It also removes any systematic intensity offsets that may be present as a result of flat background subtractions being applied as part of some acquisitions.

\begin{figure}[tbp]
\centering
\includegraphics[width=\columnwidth]{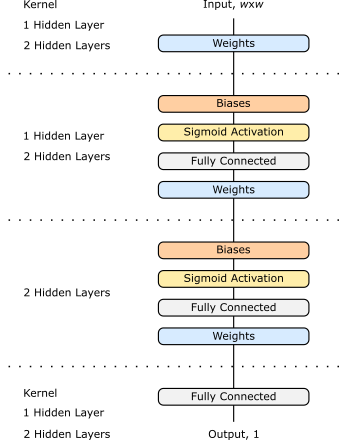}
\caption{ Architecture of kernels and MLPs with 1 and 2 hidden layers for an input size, $w$. Kernels weight the input and then fully connect it to the output. MLPs also have additional biases, sigmoid activation, a fully connected layer and weights for each hidden layer. Every weights and biases tensor has $w^2$ parameters, Fully connected layers connect to $w^2$ nodes, except for the last which connects to a single node. }
\label{kernel+MLP_architecture}
\end{figure}

\vspace{\extraspace}
\noindent \textbf{Optimization:} The ADAM solver was used with the parameters recommended in \cite{kingma2014adam} for 60000 batches. The learning rate was quadratically stepped down from $\eta_0~=~0.01$ to $\eta =(1-\textit{iter}/\textit{max\_iter})^2 \eta_0$ after every 5000 iterations.

\vspace{\extraspace}
\noindent \textbf{Batch normalization:} Batch normalization layers from \cite{ioffe2015batch} were trained with a decay rate of 0.999 and batch size 32. They are used after every convolutional layer before activation, except before and after the last.

\vspace{\extraspace}
\noindent \textbf{Activation:} Outputs of each layer are ReLU\cite{nair2010rectified} activated, except for the last layer which is unactivated.

\vspace{\extraspace}
\noindent \textbf{Initialization:} All weights were Xavier\cite{glorot2010understanding} initialized and all biases were zero initialized. The last convolutional layer has no weights or biases.

\begin{table}[tbp]
\centering
\begin{tabularx}{\columnwidth}{lYYYYY}
\multicolumn{1}{l}{TEM} &&&&& \\
\hline
              & \multicolumn{5}{c}{Input Size}    \\
       & 3 & 5  & 7   & 11     & 15       \\ \hline
Kernel           & \cmark & \cmark & \cmark & \cmark & \cmark    \\
1 Hidden Layer         & \cmark & \cmark & \cmark & \xmark & \xmark \\
2 Hidden Layers     & \xmark & \cmark & \cmark & \xmark & \xmark\\
\hline

&&&&&\\
\multicolumn{1}{l}{STEM} &&&&& \\
\hline
              & \multicolumn{5}{c}{Input Size}    \\
       & 3 & 5  & 7   & 11     & 15       \\ \hline
Kernel           & \cmark & \cmark & \cmark & \cmark & \cmark    \\
1 Hidden Layer          & \cmark & \cmark & \cmark & \xmark & \xmark \\
2 Hidden Layers     & \xmark & \xmark & \xmark & \xmark & \xmark\\
\hline

&&&&&\\
\multicolumn{1}{l}{TEM+STEM} &&&&& \\
\hline
              & \multicolumn{5}{c}{Input Size}    \\
       & 3 & 5  & 7   & 11     & 15       \\ \hline
Kernel           & \cmark & \cmark & \cmark & \cmark & \cmark    \\
1 Hidden Layer         & \cmark & \cmark & \cmark & \cmark & \xmark \\
2 Hidden Layers     & \cmark & \xmark & \cmark & \xmark & \xmark\\
\hline
\end{tabularx}
\caption{ Kernels and MLPs trained to approximate latent depth 16 TEM, STEM and TEM+STEM autoencoders. }
\label{mlps_table}
\end{table}

\section{Kernels and Multilayer Perceptrons}\label{sec_kernels+MLPs}

\begin{figure*}[tbp]
\centering
\includegraphics[width=\textwidth, trim=0cm 0.3cm 0cm .2cm, clip]{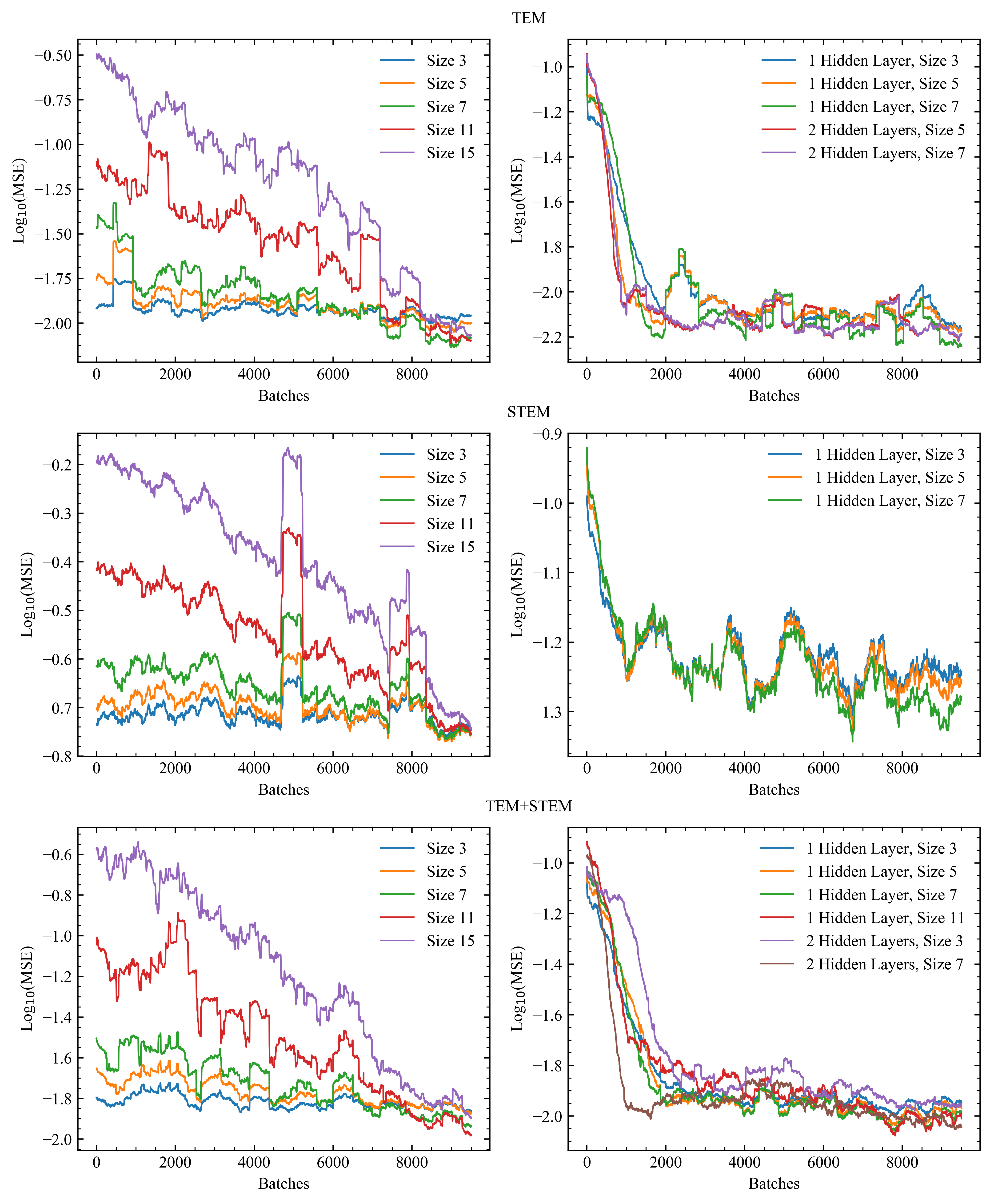}
\caption{ Kernel and MLP learning curves for TEM, STEM and TEM+STEM. Larger input sizes and more hidden layers are correlated with lower MSEs. TEM MSEs are lowest, STEM MSEs are the highest and TEM+STEM MSEs are intermediate. Similarities in  the graphs are a result of sets of kernels and MLPs being shown training data in the same order. Learning curves are moving averages of 500 iterations.}
\label{kernel+MLP_learning_curves}
\end{figure*}

\noindent Our latent depth 16 autoencoders have been used to train TEM, STEM and TEM+STEM kernels and MLPs. Many of these have higher computational efficiencies than the autoencoders. Their lower capacities also make them less susceptible to overfitting and checkerboard artefacts\cite{sugawara2018super}.

\begin{table*}[tbp]
\centering
\begin{tabularx}{\textwidth}{lYYYYYY}
\hline
              & \multicolumn{2}{c}{TEM} & \multicolumn{2}{c}{STEM} & \multicolumn{2}{c}{TEM+STEM}    \\
       & Mean & Std Dev  & Mean   & Std Dev  & Mean   & Std Dev  \\ \hline
Autoencoder, Latent Depth 1 & 0.026422 & 0.000033 & - & - & 0.275182 & 0.002885 \\
Autoencoder, Latent Depth 2 & - & - & - & - & 0.248378 & 0.003710 \\
Autoencoder, Latent Depth 4 & 0.018571 & 0.000005 & 0.387636 & 0.001898 & 0.236593 & 0.003277 \\
Autoencoder, Latent Depth 8 & - & - & - & - & 0.219717 & 0.003008 \\
Autoencoder, Latent Depth 16 & 0.014360 & \textbf{0.000004} & 0.314286 & 0.002043 & 0.212957 & 0.003008 \\
Autoencoder, Latent Depth 32 & - & - & - & - & \textbf{0.191258} & 0.001654 \\
Autoencoder, Latent Depth 64 & \textbf{0.013034} & 0.000008 & \textbf{0.219523} & \textbf{0.001777} & 0.192866 & \textbf{0.001395} \\
\hline
Kernel, Input Size 3 & 0.010822 & 0.000288 & 0.181633 & 0.002732 & 0.014224 & 0.000436 \\
Kernel, Input Size 5 & 0.009728 & 0.000398 & 0.177021 & 0.002536 & 0.014071 & 0.000518 \\
Kernel, Input Size 7 & 0.008032 & 0.000410 & 0.178340 & 0.002297 & 0.012350 & 0.000599 \\
Kernel, Input Size 11 & 0.008217 & 0.000191 & 0.181634 & 0.002163 & 0.011783 & 0.000908 \\
Kernel, Input Size 15 & 0.000929 & 0.000522 & 0.190376 & 0.003822 & 0.015055 & 0.001629 \\
MLP, 1 Hidden Layer, Input Size 3 & 0.007167 & 0.000188 & 0.057006 & \textbf{0.000905} & 0.011494 & 0.000271 \\
MLP, 1 Hidden Layer, Input Size 5 & 0.006934 & 0.000128 & 0.054407 & 0.001163 & 0.010921 & 0.000334 \\
MLP, 1 Hidden Layer, Input Size 7 & \textbf{0.005917} & \textbf{0.000113} & \textbf{0.050175} & 0.001688 & 0.010340 & 0.000308 \\
MLP, 1 Hidden Layer, Input Size 11 & - & - & - & - & 0.009965 & 0.000342 \\
MLP, 2 Hidden Layers, Input Size 3 & - & - & - & - & 0.011027 & \textbf{0.000095} \\
MLP, 2 Hidden Layers, Input Size 5 & 0.006590 & 0.000263 & - & - & - & - \\
MLP, 2 Hidden Layers, Input Size 7 & 0.006614 & 0.000247 & - & - & \textbf{0.009252} & 0.000292 \\
\hline
\end{tabularx}
\caption{ Means and standard deviations of autoencoder, kernel and MLP MSEs in their last 500 training batches for TEM, STEM and TEM+STEM. The highest performing autoencoders and kernel or MLP are emboldened for each column.}
\label{MSE_performance}
\end{table*}

\subsection{Architecture}
\noindent Our kernel and MLP architecture is shown in fig.~\ref{kernel+MLP_architecture}. Kernels weight their input then fully connect it to their outputs. MLPs also have additional biases, sigmoid activation, a fully connected layer and weights for each hidden layer. For a $w$$\times$$w$ input, every weights and biases tensor has $w^2$ parameters and hidden layers have $w^2$ nodes. Our fully trained kernels and MLPs are summarized in table~\ref{mlps_table}.

\subsection{Learning Policy}
\noindent In this subsection, we present our kernel and MLP training hyperparameters and learning protocol for the learning curves shown in fig.~\ref{kernel+MLP_learning_curves}.

\vspace{\extraspace}
\noindent \textbf{Batched training:} Sets of kernels and MLPs were trained together and shown training data in the same order. This is why some learning curves look similar. The sets were: TEM kernels, TEM MLPs with 1 hidden layer, TEM MLPs with 2 hidden layers, STEM kernels, STEM MLPs with 1 hidden layer, STEM MLPs with 2 hidden layers, TEM+STEM kernels, TEM+STEM MLPs with 1 hidden layer and TEM+STEM MLPs with 2 hidden layers.

\vspace{\extraspace}
\noindent \textbf{Loss metric:} Our kernels and MLPs were trained to minimize MSEs between single $d$$\times$$d$ crops from TEM, STEM or TEM+STEM images and the respective outputs from latent depth 16 TEM, STEM and TEM+STEM autoencoders. Crop sizes were at least $d=w_\mathrm{max}+5$, where $w_\mathrm{max}$ is the largest input size, $w$, in a set of kernels or autoencoders being trained together. To avoid edge artifacts, MSEs were calculated using pixels at least $(w~-~1)/2$ from the edges of the crops.

\vspace{\extraspace}
\noindent \textbf{Optimization:} The ADAM solver was used with the parameters recommended in \cite{kingma2014adam} for 10000 batches. The learning rate was quadratically stepped down from $\eta_0~=~0.01$ to $\eta =(1-\textit{iter}/\textit{max\_iter})^2 \eta_0$ after each iteration.

\vspace{\extraspace}
\noindent \textbf{Activation:} All neurons are sigmoid activated.

\vspace{\extraspace}
\noindent \textbf{Initialization:} All weights were Xavier\cite{glorot2010understanding} initialized. All biases were zero initialized. The last convolutional layer has no weights or biases.

\section{Performance}\label{sec_performance}

\noindent Autoencoder, kernel and MLP MSE performances for TEM, STEM and TEM+STEM training are summarized in table~\ref{MSE_performance} for comparison. They are for the last 500 training batches of the learning curves in fig.~\ref{autoenc_learning_curves} and fig.~\ref{kernel+MLP_learning_curves}. The performance of each autoencoder, kernel and MLP for the domain it was trained for is exemplified by figs~\ref{examples1}-\ref{examples8} in the appendix.

Autoencoder MSE means and standard deviations are negatively correlated with latent depth and are lowest for either the highest or second highest latent depths we trained. This is a result of larger latent spaces encoding more information, making decoding easier.

Kernel and MLP MSE means and standard deviations are negatively correlated with input sizes and the number of hidden layers (a kernel is a 0 hidden layer MLP). This is a result of kernels and MLPs with higher input sizes and more hidden layers having more trainable parameters. Since their successive layers are fully connected, they have higher capacities and were therefore able to more accurately approximate the 4$\times$ compression autoencoders.

STEM MSEs are systematically higher than TEM MSEs, with TEM+STEM MSEs inbetween. In part, this is because local variations in intensity are much higher in STEM images, making them harder to autoencode. This meant that STEM autoencoder learning curves did not plateau in our 60000 batch training schedule, resulting in STEM MSEs being above their convergence limit. Nevertheless, this has not had a large qualitative effect on the ability of kernels and MLPs to restore STEM images, as shown by the examples in the appendix.

\section{Summary}

\begin{itemize}
\item We have presented TEM, STEM and TEM+STEM denoising autoencoders for a range of compression ratios between 1$\times$ and 64$\times$.
\item Kernels and MLPs with input sizes between 3 and 15 have been trained approximate the restorations of 4$\times$ compression TEM, STEM and TEM+STEM autoencoders.
\item Fully trained autoencoders, kernels and MLPs have been made publicly available with example usage: \url{https://github.com/Jeffrey-Ede/Denoising-Kernels-MLPs-Autoencoders}.
\item Our autoencoder, kernel and MLP architectures, training hyperparameters and learning protocols are detailed.
\item Example applications of all our autoencoders, kernels and MLPs are provided in the appendix. Some kernels are also presented.
\end{itemize}

\bibliographystyle{ieeetr}
\bibliography{bibliography}

\clearpage

\onecolumn
\section{Appendix}
\noindent In this appendix, we show example applications of each of our kernels, MLPs and autoencoders to the domains they were trained for. These are followed by our 3$\times$3, 5$\times$5, 7$\times$7 and 11$\times$11 kernels for TEM, STEM and TEM+STEM.


\vspace{0.3cm}
\begin{tabular*}{\linewidth}{l}
\multicolumn{1}{l}{TEM}\\
\hline\\
\end{tabular*}

\begin{equation*}
\left( \begin{array}{ccc}
0.064 & 0.135 & 0.064 \\
0.135 & 0.218 & 0.135 \\
0.064 & 0.135 & 0.064 \end{array} \right)
\end{equation*}

\begin{equation*}
\left( \begin{array}{ccccc}
-0.084 & -0.001& 0.030 & -0.001 & -0.084 \\
-0.001 & 0.107 & 0.157 & 0.107 & -0.001 \\
0.030 & 0.157 & 0.220 & 0.157 & 0.030 \\
-0.001 & 0.107 & 0.157 & 0.107 & -0.001 \\
-0.084 & -0.001 & 0.030 & -0.001 & -0.084 \end{array} \right)
\end{equation*}

\begin{equation*}
\left( \begin{array}{ccccccc}
-0.039 & -0.038 & -0.025 & -0.018 & -0.025 & -0.038 & -0.039\\
-0.038 & -0.009 & 0.035 & 0.058 & 0.035 & -0.009 & -0.038\\
-0.025 & 0.035 & 0.114 & 0.153 & 0.114 & 0.035 & -0.025\\
-0.018 & 0.058 & 0.153 & 0.204 & 0.153 & 0.058 & -0.018\\
-0.025 & 0.035 & 0.114 & 0.153 & 0.114 & 0.035 & -0.025\\
-0.038 & -0.009 & 0.035 & 0.058 & 0.035 & -0.009 & -0.038\\
-0.039 & -0.038 & -0.025 & -0.018 & -0.025 & -0.038 & -0.039\end{array} \right)
\end{equation*}

\begin{equation*}
\left( \begin{array}{ccccccccccc}
-0.017 & 0.000 & 0.001 & -0.005 & -0.006 & -0.006 & -0.006 & -0.005 & 0.001 & 0.000 & -0.017\\
0.000 & 0.009 & -0.001 & -0.013 & -0.015 & -0.014 & -0.015 & -0.013 & -0.001 & 0.009 & 0.000\\
0.001 & -0.001 & -0.015 & -0.022 & -0.010 & -0.001 & -0.010 & -0.022 & -0.015 & -0.001 & 0.001\\
-0.005 & -0.013 & -0.022 & -0.007 & 0.035 & 0.059 & 0.035 & -0.007 & -0.022 & -0.013 & -0.005\\
-0.006 & -0.015 & -0.010 & 0.035 & 0.111 & 0.152 & 0.111 & 0.035 & -0.010 & -0.015 & -0.006\\
-0.006 & -0.014 & -0.001 & 0.059 & 0.152 & 0.202 & 0.152 & 0.059 & -0.001 & -0.014 & -0.006\\
-0.006 & -0.015 & -0.010 & 0.035 & 0.111 & 0.152 & 0.111 & 0.035 & -0.010 & -0.015 & -0.006\\
-0.005 & -0.013 & -0.022 & -0.007 & 0.035 & 0.059 & 0.035 & -0.007 & -0.022 & -0.013 & -0.005\\
0.001 & -0.001 & -0.015 & -0.022 & -0.010 & -0.001 & -0.010 & -0.022 & -0.015 & -0.001 & 0.001\\
0.000 & 0.009 & -0.001 & -0.013 & -0.015 & -0.014 & -0.015 & -0.013 & -0.001 & 0.009 & 0.000\\
-0.017 & 0.000 & 0.001 & -0.005 & -0.006 & -0.006 & -0.006 & -0.005 & 0.001 & 0.000 & -0.017\end{array} \right)
\end{equation*}

\begin{tabular*}{\linewidth}{l}
\multicolumn{1}{l}{STEM}\\
\hline\\
\end{tabular*}

\begin{equation*}
\left( \begin{array}{ccc}
0.108 & 0.111 & 0.108 \\
0.111 & 0.109 & 0.111 \\
0.108 & 0.111 & 0.108 \end{array} \right)
\end{equation*}

\begin{equation*}
\left( \begin{array}{ccccc}
0.004 & 0.026 & 0.040 & 0.026 & 0.004 \\
0.026 & 0.057 & 0.089 & 0.057 & 0.026 \\
0.040 & 0.089 & 0.089 & 0.089 & 0.040 \\
0.026 & 0.057 & 0.089 & 0.057 & 0.026 \\
0.004 & 0.026 & 0.040 & 0.026 & 0.004 \end{array} \right)
\end{equation*}

\begin{equation*}
\left( \begin{array}{ccccccc}
-0.016 & -0.004 & 0.007 & 0.012 & 0.007 & -0.004 & -0.016\\
-0.004 & 0.007 & 0.026 & 0.035 & 0.026 & 0.007 & -0.004\\
0.007 & 0.026 & 0.057 & 0.071 & 0.057 & 0.026 & 0.007\\
0.012 & 0.035 & 0.071 & 0.089 & 0.071 & 0.035 & 0.012\\
0.007 & 0.026 & 0.057 & 0.071 & 0.057 & 0.026 & 0.007\\
-0.004 & 0.007 & 0.026 & 0.035 & 0.026 & 0.007 & -0.004\\
-0.016 & -0.004 & 0.007 & 0.012 & 0.007 & -0.004 & -0.016\end{array} \right)
\end{equation*}

\begin{equation*}
\left( \begin{array}{ccccccccccc}
0.012 & 0.003 & -0.001 & -0.002 & 0.000 & 0.002 & 0.000 & -0.002 & -0.001 & 0.003 & 0.012\\
0.003 & -0.006 & -0.009 & -0.007 & -0.001 & 0.005 & -0.001 & -0.007 & -0.009 & -0.006 & 0.003\\
-0.001 & -0.009 & -0.010 & -0.001 & 0.007 & 0.013 & 0.007 & -0.001 & -0.010 & -0.009 & -0.001\\
-0.002 & -0.007 & -0.001 & 0.012 & 0.029 & 0.038 & 0.029 & 0.012 & -0.001 & -0.007 & -0.002\\
0.000 & -0.001 & 0.007 & 0.029 & 0.055 & 0.070 & 0.055 & 0.029 & 0.007 & -0.001 & 0.000\\
0.002 & 0.005 & 0.013 & 0.038 & 0.070 & 0.089 & 0.070 & 0.038 & 0.013 & 0.005 & 0.002\\
0.000 & -0.001 & 0.007 & 0.029 & 0.055 & 0.070 & 0.055 & 0.029 & 0.007 & -0.001 & 0.000\\
-0.002 & -0.007 & -0.001 & 0.012 & 0.029 & 0.038 & 0.029 & 0.012 & -0.001 & -0.007 & -0.002\\
-0.001 & -0.009 & -0.010 & -0.001 & 0.007 & 0.013 & 0.007 & -0.001 & -0.010 & -0.009 & -0.001\\
0.003 & -0.006 & -0.009 & -0.007 & -0.001 & 0.005 & -0.001 & -0.007 & -0.009 & -0.006 & 0.003\\
0.012 & 0.003 & -0.001 & -0.002 & 0.000 & 0.002 & 0.000 & -0.002 & -0.001 & 0.003 & 0.012\end{array} \right)
\end{equation*}

\begin{tabular*}{\linewidth}{l}
\multicolumn{1}{l}{TEM+STEM}\\
\hline\\
\end{tabular*}

\begin{equation*}
\left( \begin{array}{ccc}
0.093 & 0.124 & 0.093 \\
0.124 & 0.149 & 0.124 \\
0.093 & 0.124 & 0.093 \end{array} \right)
\end{equation*}

\begin{equation*}
\left( \begin{array}{ccccc}
-0.061 & 0.016 & 0.042 & 0.016 & -0.061 \\
0.016 & 0.091 & 0.116 & 0.091 & 0.016 \\
0.042 & 0.116 & 0.142 & 0.116 & 0.042 \\
0.016 & 0.091 & 0.116 & 0.091 & 0.016 \\
-0.061 & 0.016 & 0.042 & 0.016 & -0.061 \end{array} \right)
\end{equation*}

\begin{equation*}
\left( \begin{array}{ccccccc}
-0.077 & -0.037 & -0.008 & 0.001 & -0.008 & -0.037 & -0.077\\
-0.037 & 0.016 & 0.052 & 0.063 & 0.052 & 0.016 & -0.037\\
-0.008 & 0.052 & 0.095 & 0.110 & 0.095 & 0.052 & -0.008\\
0.001 & 0.063 & 0.110 & 0.127 & 0.110 & 0.063 & 0.001\\
-0.008 & 0.052 & 0.095 & 0.110 & 0.095 & 0.052 & -0.008\\
-0.037 & 0.016 & 0.052 & 0.063 & 0.052 & 0.016 & -0.037\\
-0.077 & -0.037 & -0.008 & 0.001 & -0.008 & -0.037 & -0.077\end{array} \right)
\end{equation*}

\begin{equation*}
\left( \begin{array}{ccccccccccc}
0.005 & -0.003 & -0.015 & -0.022 & -0.019 & 0.017 & -0.019 & -0.022 & -0.015 & -0.003 & 0.005\\
-0.003 & -0.008 & -0.013 & -0.013 & -0.004 & 0.001 & -0.004 & -0.013 & -0.013 & -0.008 & 1\\
-0.015 & -0.013 & -0.011 & -0.001 & 0.017 & 0.025 & 0.017 & -0.001 & -0.011 & -0.013 & -0.015\\
-0.022 & -0.013 & -0.001 & 0.021 & 0.050 & 0.062 & 0.050 & 0.021 & -0.001 & -0.013 & -0.022\\
-0.019 & -0.004 & 0.017 & 0.050 & 0.088 & 0.105 & 0.088 & 0.050 & 0.017 & -0.004 & -0.019\\
0.017 & 0.001 & 0.025 & 0.062 & 0.105 & 0.123 & 0.105 & 0.062 & 0.025 & 0.001 & 0.017\\
-0.019 & -0.004 & 0.017 & 0.050 & 0.088 & 0.105 & 0.088 & 0.050 & 0.017 & -0.004 & -0.019\\
-0.022 & -0.013 & -0.001 & 0.021 & 0.050 & 0.062 & 0.050 & 0.021 & -0.001 & -0.013 & -0.022\\
-0.015 & -0.013 & -0.011 & -0.001 & 0.017 & 0.025 & 0.017 & -0.001 & -0.011 & -0.013 & -0.015\\
-0.003 & -0.008 & -0.013 & -0.013 & -0.004 & 0.001 & -0.004 & -0.013 & -0.013 & -0.008 & -0.003\\
0.005 & -0.003 & -0.015 & -0.022 & -0.019 & 0.017 & -0.019 & -0.022 & -0.015 & -0.003 & 0.005\end{array} \right)
\end{equation*}



\clearpage


\begin{figure*}[tbp]
\centering
\includegraphics[width=\textwidth]{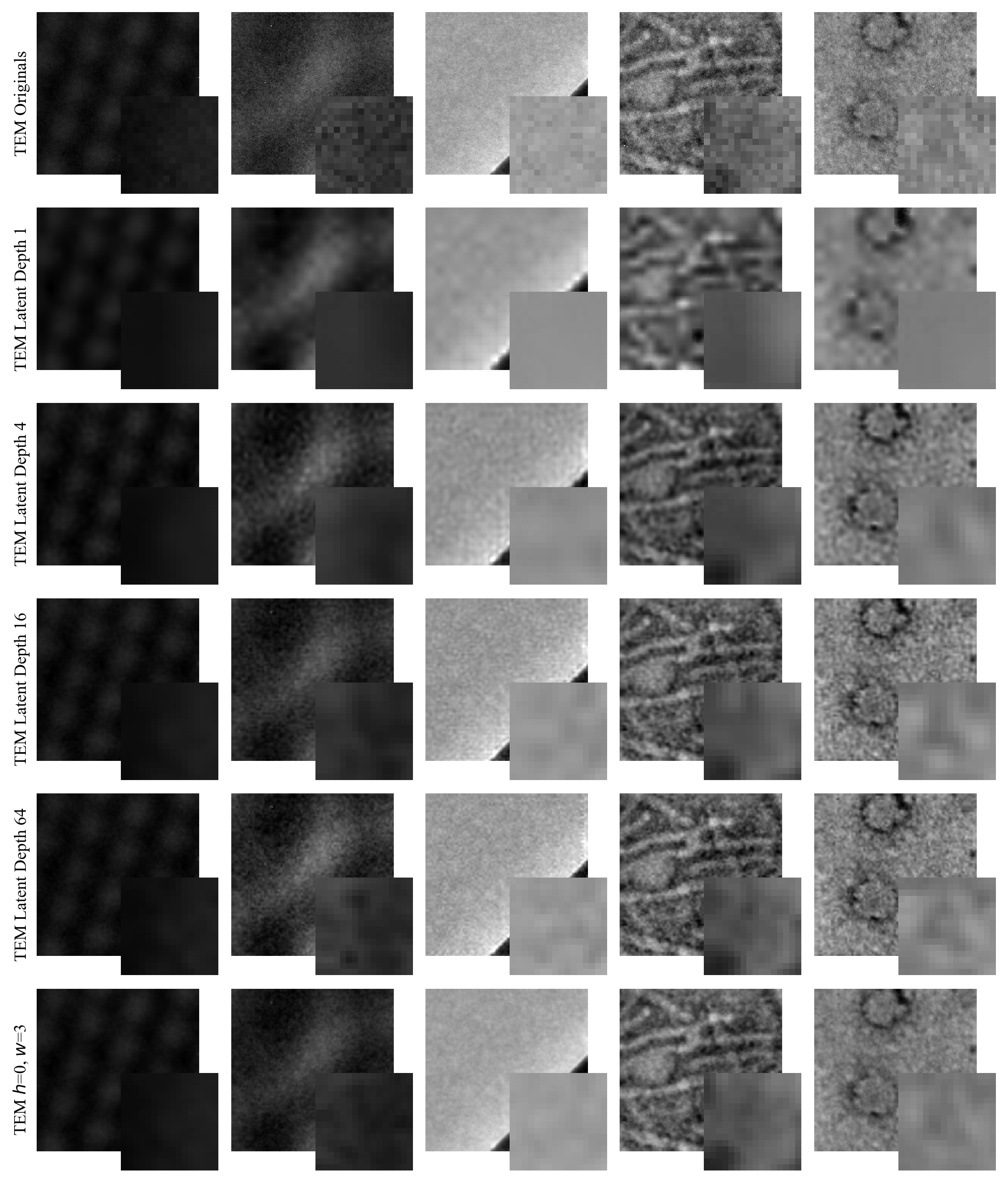}
\caption{ \finalcapttxtstart }
\label{examples1}
\end{figure*}

\begin{figure*}[tbp]
\centering
\includegraphics[width=\textwidth]{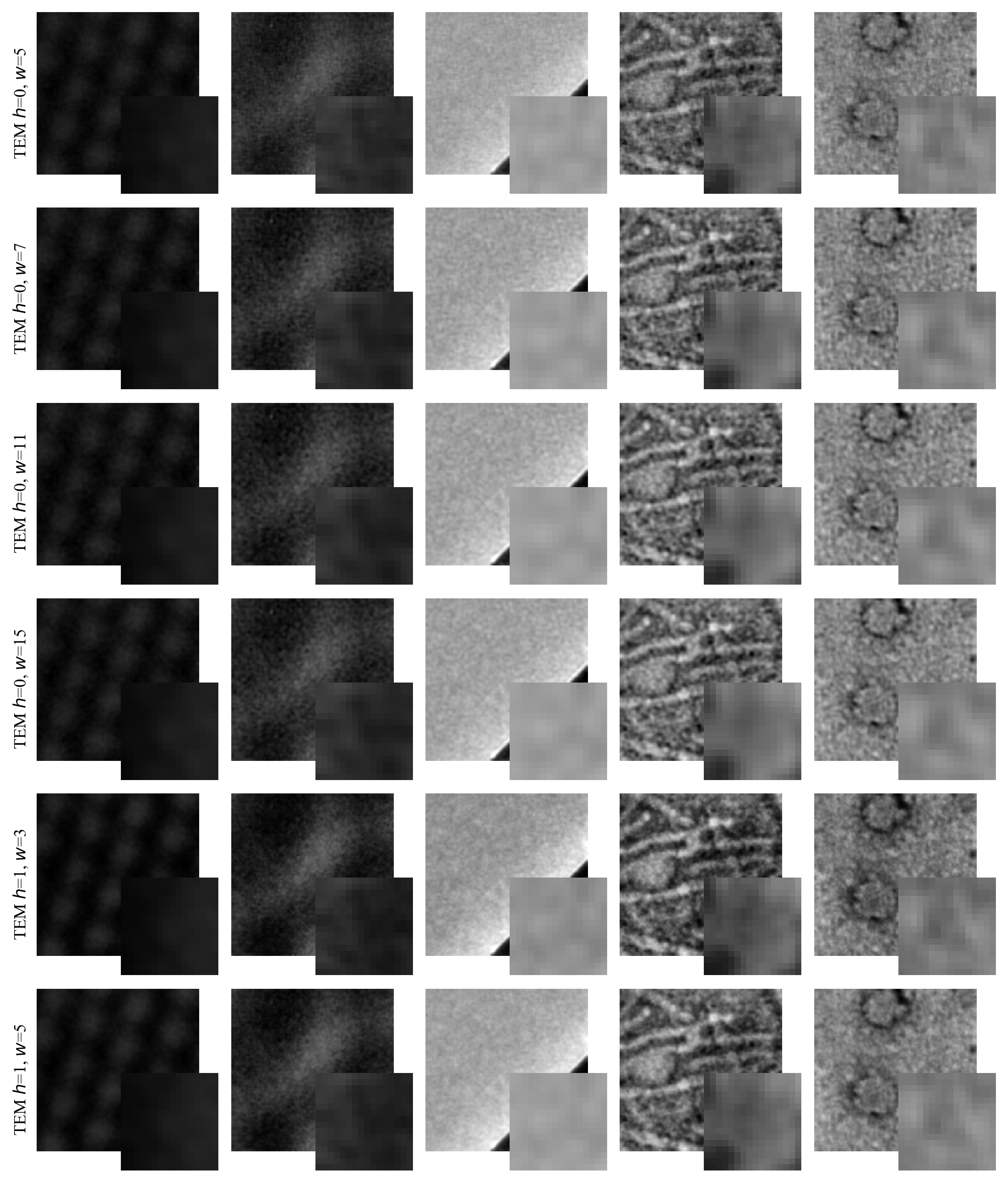}
\caption{ \finalcapttxt}
\label{examples2}
\end{figure*}

\begin{figure*}[tbp]
\centering
\includegraphics[width=\textwidth]{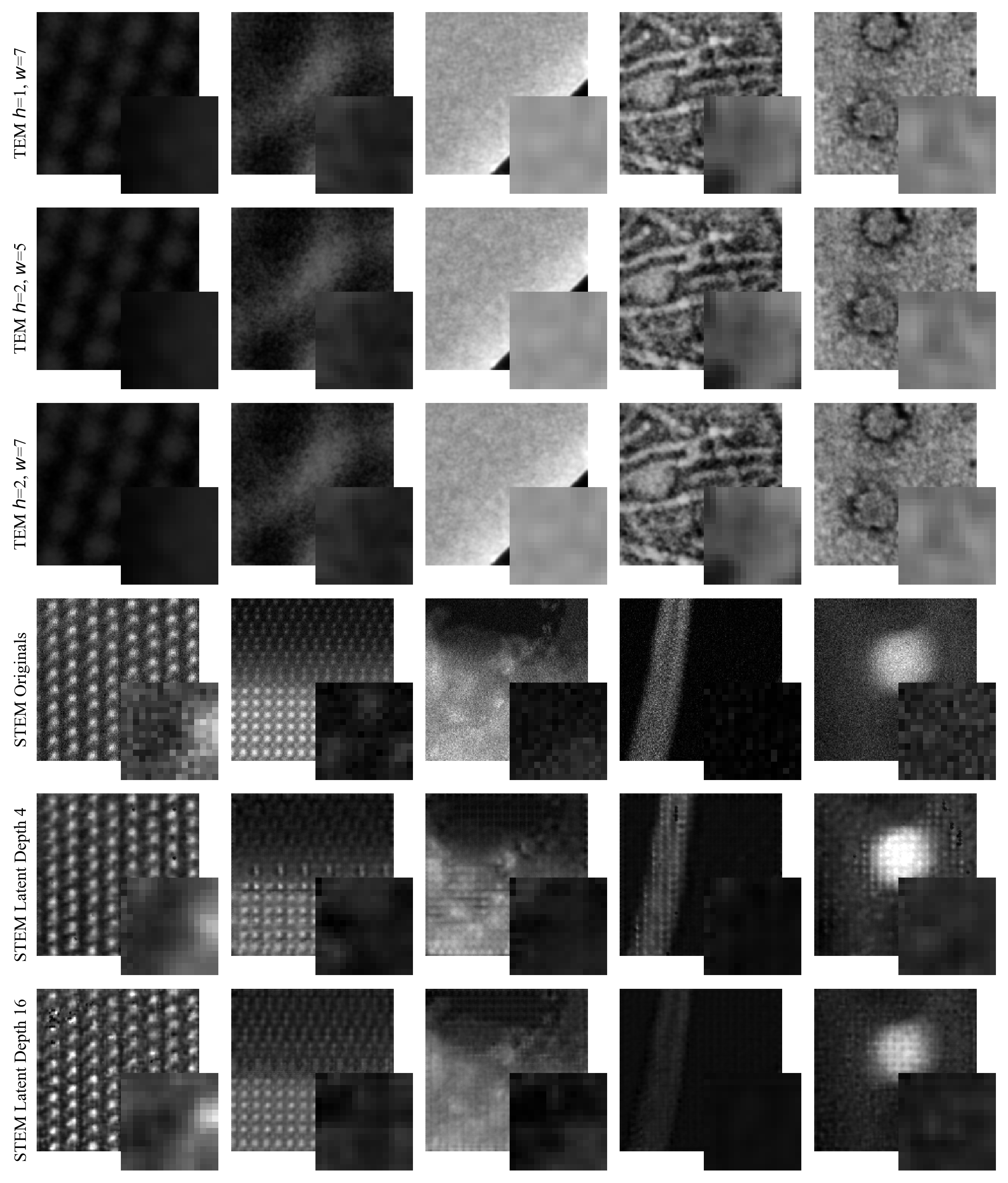}
\caption{ \finalcapttxt }
\label{examples3}
\end{figure*}

\begin{figure*}[tbp]
\centering
\includegraphics[width=\textwidth]{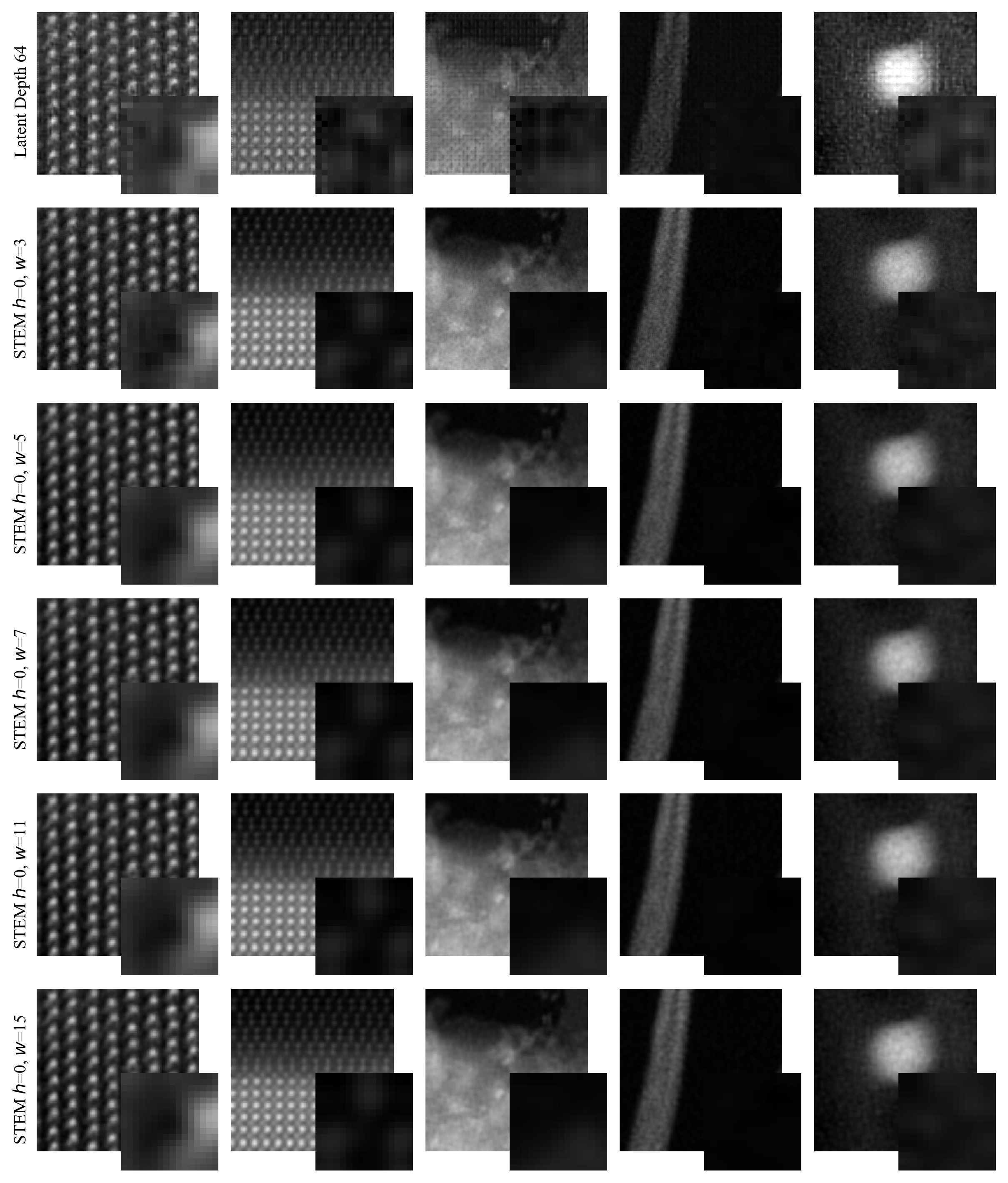}
\caption{ \finalcapttxt}
\label{examples4}
\end{figure*}

\begin{figure*}[tbp]
\centering
\includegraphics[width=\textwidth]{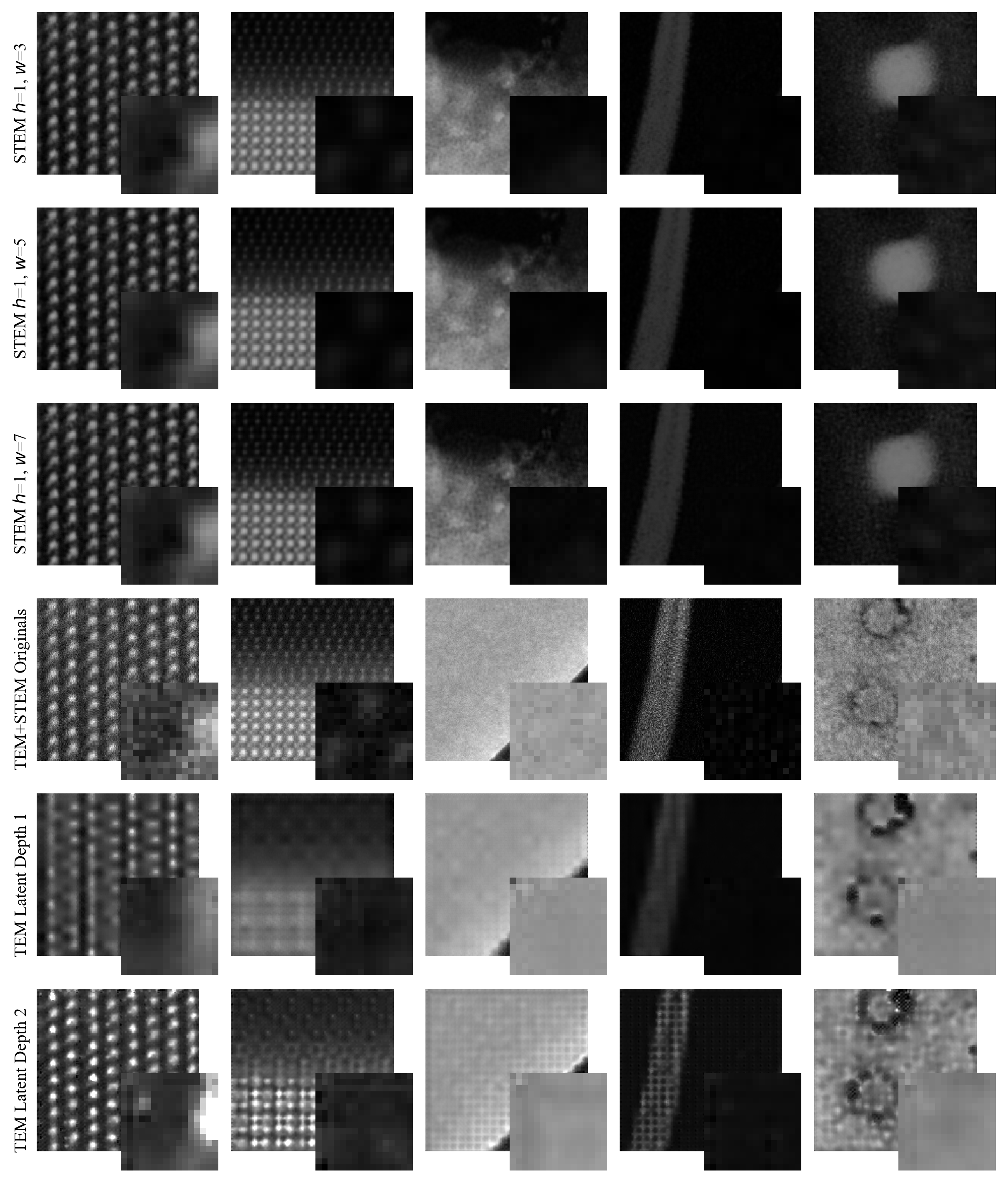}
\caption{ \finalcapttxt }
\label{examples5}
\end{figure*}

\begin{figure*}[tbp]
\centering
\includegraphics[width=\textwidth]{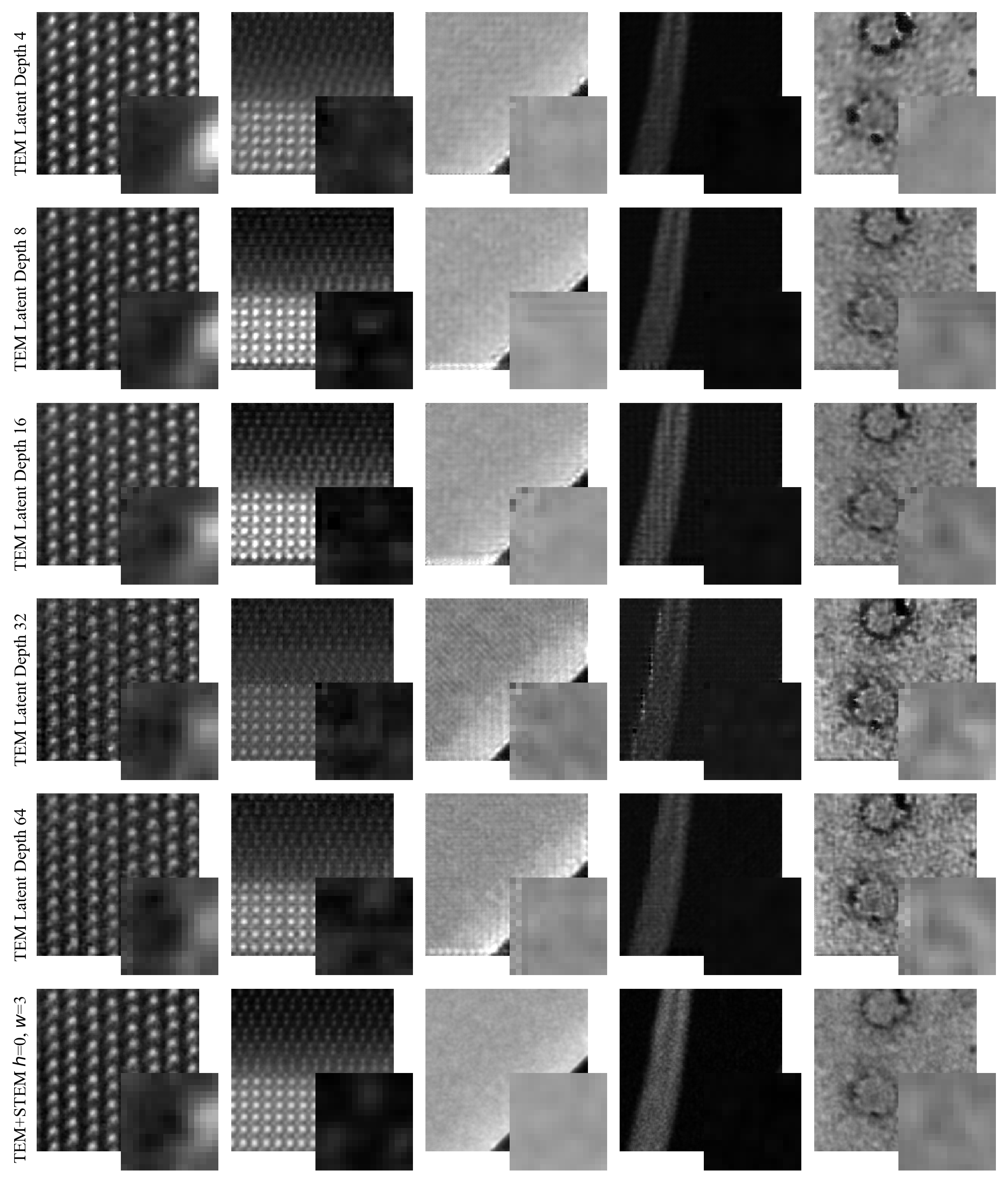}
\caption{ \finalcapttxt}
\label{examples6}
\end{figure*}

\begin{figure*}[tbp]
\centering
\includegraphics[width=\textwidth]{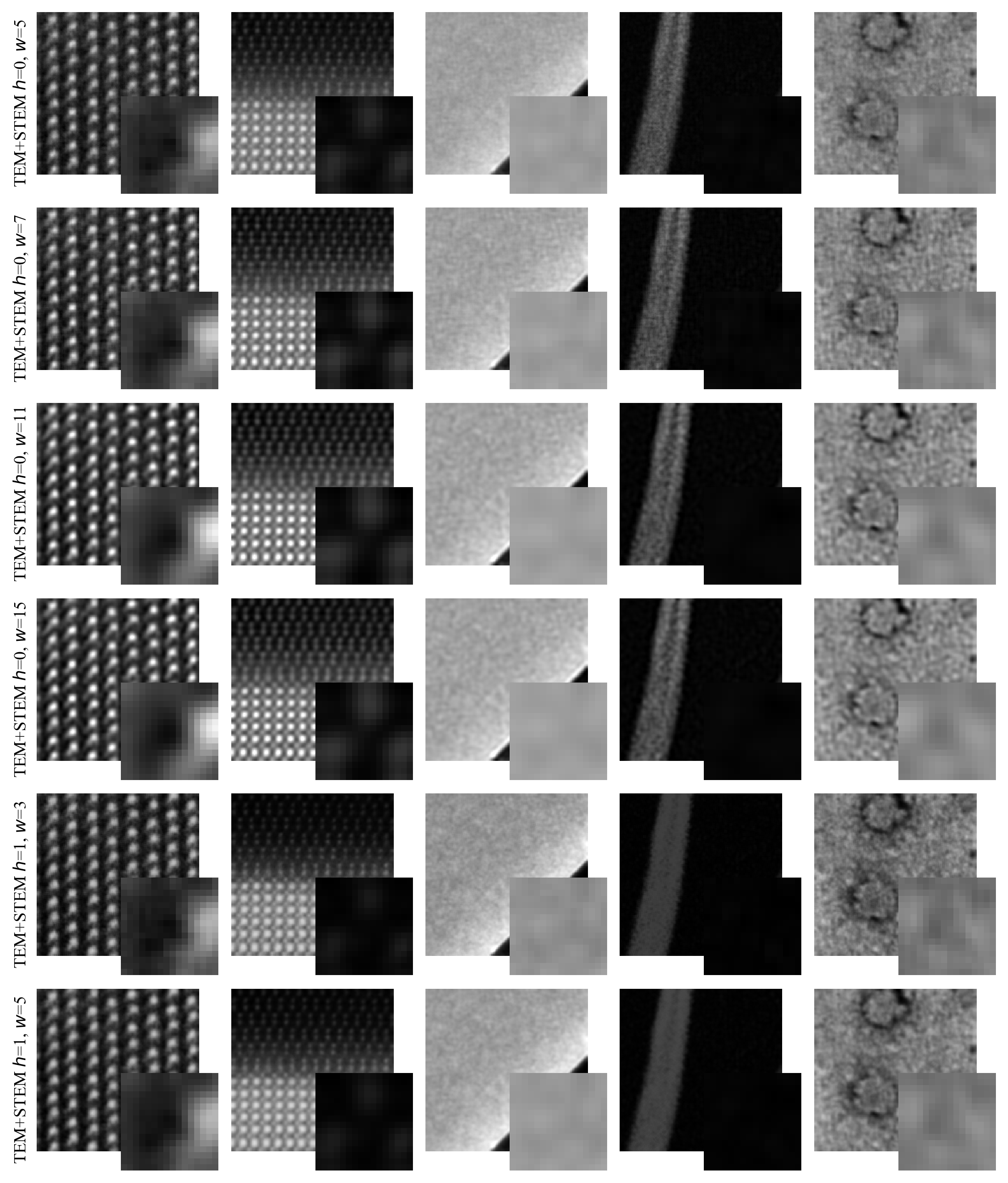}
\caption{ \finalcapttxt }
\label{examples7}
\end{figure*}

\begin{figure*}[tbp]
\centering
\includegraphics[width=\textwidth]{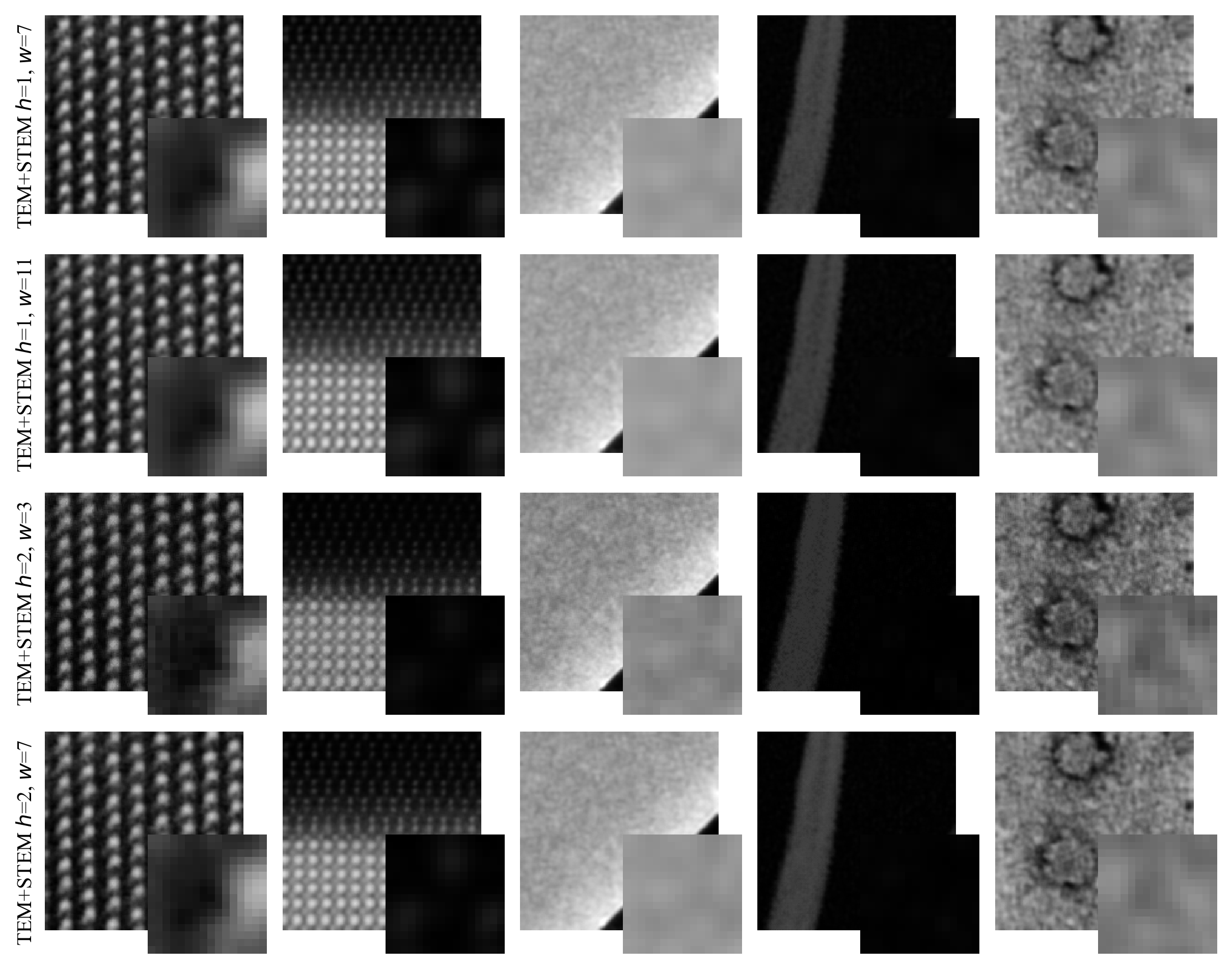}
\caption{ \finalcapttxt }
\label{examples8}
\end{figure*}
\end{document}